\def\year{2020}\relax
\documentclass[letterpaper]{article} 
\usepackage{aaai20}  
\usepackage{times}  
\usepackage{helvet} 
\usepackage{courier}  
\usepackage[hyphens]{url}  
\usepackage{graphicx} 
\usepackage{graphicx}  
\title{\mname: Deep Recurrent Inverse TreatmEnt Weighting for Adjusting Time-varying Confounding in Modern Longitudinal Observational Data}

\newcommand{\mname}{\texttt{DeepRite}\xspace}
\newcommand{\mnameS}{\texttt{DeepRite-S}\xspace}

\author{Yanbo Xu\textsuperscript{\rm 1}, Cao Xiao\textsuperscript{\rm 2}, Jimeng Sun\textsuperscript{\rm 3}\\ 
\textsuperscript{\rm 1} Georgia Institute of Technology, \textsuperscript{\rm 2} IQVIA, \textsuperscript{\rm 3} University of Illinois Urbana-Champaign\\ 
yxu465@gatech.edu, cao.xiao@iqvia.com, jimeng@illinois.edu 
}

\usepackage{graphicx} 
\usepackage{amsmath,bm}
\usepackage{amsfonts}
\usepackage[hidelinks]{hyperref}
\usepackage{subcaption}
\usepackage{amssymb}
\usepackage{framed}
\usepackage{stfloats}
\usepackage{xspace}

\usepackage[utf8]{inputenc} 
\usepackage{url}            
\usepackage{booktabs}       
\usepackage{nicefrac}       
\usepackage{microtype}      

\usepackage{xcolor}

\usepackage{enumitem}
\usepackage{multirow}
\usepackage[algo2e, linesnumbered,ruled, vlined]{algorithm2e}
\usepackage{wrapfig}

\usepackage{color, colortbl}
\definecolor{Gray}{gray}{0.9}

\SetKwInput{KwInput}{Input}                
\SetKwInput{KwOutput}{Output}              

\SetCommentSty{mycommfont}

\newcommand{\E}[0]{\mathbb{E}}

\newcommand{\Prob}{\text{Pr}}

\newcommand{\bh}{\textbf{h}}

\newcommand{\vx}{\overline{x}}

\newcommand{\va}{\overline{a}}

\newcommand{\vX}{\overline{X}}

\newcommand{\vA}{\overline{A}}

\newcommand{\Loss}{\mathbb{L}}
\newcommand{\A}{\mathcal{A}}
\newcommand{\X}{\mathcal{X}}
\newcommand{\Y}{\mathcal{Y}}

\newcommand{\B}{\mathcal{B}}

\newcommand{\dataD}{\mathbb{D}}

\begin{document}

\maketitle
\begin{abstract}

Counterfactual prediction is about predicting  outcome of the unobserved situation from the data. For example, given patient is on drug A, what would be the outcome if she switch to drug B. 
Most of existing works focus on modeling counterfactual outcome based on static data. However, many applications have time-varying confounding effects such as multiple treatments over time. How to model such time-varying effects from longitudinal observational data? How to model complex high-dimensional dependency in the data? 
To address these challenges, we propose Deep Recurrent Inverse TreatmEnt  weighting (\mname) by incorporating recurrent neural networks into two-phase adjustments for the existence of time-varying confounding in modern longitudinal data. In phase I {\it cohort reweighting} we fit one network for emitting time dependent inverse probabilities of treatment, use them to generate a pseudo balanced cohort. In phase II {\it outcome progression}, we input the adjusted data to the subsequent predictive network for making counterfactual predictions. We evaluate \mname on both synthetic data and a real data collected from sepsis patients in the intensive care units. \mname is shown to recover the ground truth from synthetic data, and estimate unbiased treatment effects from real data that can be better aligned with the standard guidelines for management of sepsis thanks to its applicability to create balanced cohorts. 
\end{abstract}


\section{Introduction}
\label{sec:intro}


Counterfactual predictions based on observational data is an important problem especially in medicine. The problem becomes particularly challenging when treatments can be repeated multiple times and their impacts on counterfactual predictions are confounded by past treatments and variables changing over time; this is known as the {\bf time-varying confounding problem}. 
A number of statistical methods have been proposed for adjusting time-varying confounding. They mainly fall into three categories. The first category uses \textit{inverse probability treatment weighting} (IPTW) \cite{robins2000marginal,van2007causal} to re-weight the observed data and formulate a pseudo-population mimicking the randomized study. Although Matching \cite{roy2017flame} is also widely used for balancing data in static settings, it is rarely used and hardly extended to longitudinal settings. The second category uses \textit{g-formula} to directly model the outcome progressions and simulate all potential outcomes from a pseudo-population that assumes treatments are uniformly assigned. Linear regression \cite{daniel2011gformula},  Gaussian process regression \cite{xu2016bayesian,schulam2017reliable} have been applied here. The third category is the \textit{doubly robust} method \cite{van2006targeted}, combining the above two methods so that it requires only one correctly specified model for either IPT estimations or outcome progressions. However, none of these traditional casual approaches can handle the massive amount of complex data like the high-dimensional continuously monitoring data motivated us in this paper.
With the rapid growth in dimensionality and complexity of observational data, machine learning especially deep learning (DL) have been widely used in factual predictions. The key advantage of DL models are their abilities to extract effective features and temporal patterns that co-occur frequently with a certain prediction outcome. DL has demonstrated the state-of-the-art performance in many predictive tasks, such as medical concepts construction~\cite{choi2016medical,choi2016using}, disease prediction \cite{xu2018raim,esteban2016predicting}, patient subtyping \cite{miotto2016deep} in the context of healthcare applications.
Several DL models have also been adapted in counterfactual inference. These methods include using generative adversarial networks (GAN) \cite{yoon2018ganite} to directly estimate the possible potential outcomes, and convolutional neural networks \cite{shalit2017estimating,johansson2016learning} or autoencoders \cite{atan2018deep} to learn latent representations that can balance the distributions of counterfactuals and factuals. However, the above methods are mainly designed for static setting and cannot be easily extended to longitudinal studies. Moreover, most of them only consider binary treatment assignments, and cannot be easily generalized to multiple treatments. To our best knowledge, there is no existing DL solutions in high-dimensional time-varying adjustments in longitudinal studies.

\begin{figure}[t]
    \centering
    \includegraphics[width= .49\textwidth]{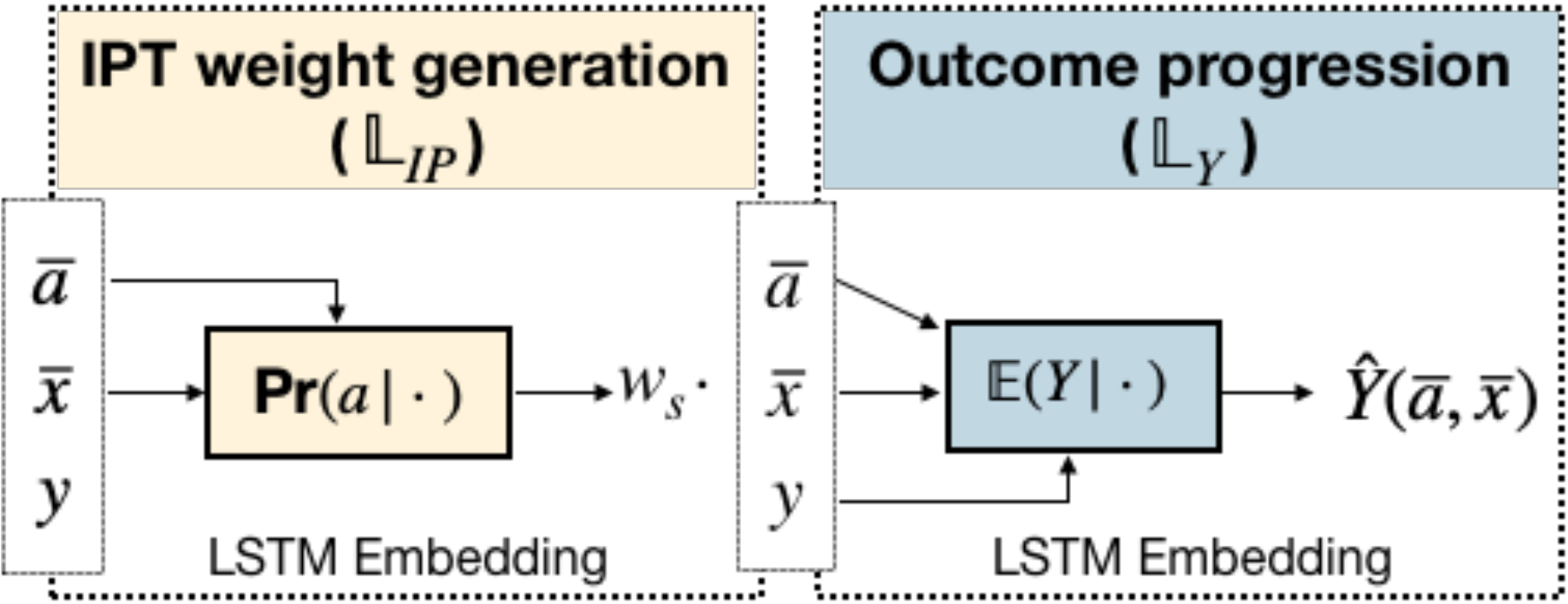}
    \caption{\mname: A generic pipeline of using recurrent inverse treatment weighting for adjusting time-varying confounding in observational longitudinal data.}
    \label{fig:framework}
\end{figure}

To bridge the gap, we propose  \mname, the Deep Recurrent Inverse TreatmEnt weighting method shown in Figure \ref{fig:framework}, for adjusting the aforementioned time-varying bias and making counterfactual predictions based on longitudinal observational data. The idea is to train two recurrent neural networks (RNNs) each encoding the complex longitudinal history respectively for emitting the time-varying IPT weights and fitting outcome progressions on the re-weighted balanced pseudo cohort. Our method falls into the third category of doubly robust adjustment in causal inference.

Overall, \mname has the following key contributions:
1) We propose a generic pipeline of using deep neural networks to remove bias that exist in longitudinal observational data for counterfactual predictions. 2) We bridge the gap and generalize the existing static DL-based counterfactual models into longitudinal studies in the presence of time-varying confounding effects. 3) By working with a challenging real-world problem, we are able to demonstrate \mname can not only obtain accurate factual predictions as the powerful DL predictive networks but also produce less biased estimations of longitudinal effects in terms of being better aligned with the standard treatment guidelines for sepsis patients in the population, as well as the practical judgements of risks for different pre-specified groups of patients.  


\section{Approach}
\noindent \textbf{Notations.} Let $\A$ be the set of $k$ treatments of our interest which can be repeatedly given to a patient, $\B$ the baseline feature space, $\X$ the time-dependent feature space and $\Y$ the set of possible outcomes. For example, given an initial baseline variable $B \in \B$, one of the $k$ treatments or no treatments may be initiated -- denoted as $A_0 \in \A \bigcup \varnothing$, the time-dependent covariates are subsequently obtained as $X_1$ and a new action (give one treatment in $A$ or not) is then assigned at the next time step as $A_1$. Following the Rubin-Neyman's causal framework \cite{rubin1974estimating,rubin2005causal}, we say by time $t$, for baseline $B$ and a sequence of covariates $\vX = (X_1, ... , X_t)$, there are a sequence of treatment assignments $\vA = (A_0, A_1,..., A_t)$ and final potential outcomes $Y\big(\vA, B, \vX \big)$\footnote{Note that we use uppercase such as $X$ to denote a variable, lowercase $x$ to denote an instance or an observation of the variable, calligraphy like $\X$ to denote the domain, overline $\vx$ to denote a sequence, and bold $\mathbf{x}$ to denote a high-dimensional vector.}. Table \ref{tab:symbol} summarizes the notations we use in this paper.
\begin{table}
\centering
\caption{Notations in this paper}
\resizebox{\columnwidth}{!}{
\begin{tabular}{l l}
\toprule[1pt]
 \bf Notation & \bf Definition\\
  \midrule
$\A$ & The set of $k$ treatments of interest \\ 
$\B$ & The domain of baseline features \\
$\X$ & The domain of time-varying features \\ 
$\Y$  & The domain of outcomes\\
$A$ / $a$ & A variable / an instance of treatment\\
$\vA$ / $\va$ & A sequence / sequential instances of $A$\\
$B$ / $b$ & A variable / an instance of baseline confounder\\
$X$ / $x$ & A variable / an instance of time-varying confounder\\
$\vX$ / $\vx$  & A sequence / sequential instances of $X$\\
$Y$ / $y$ & A variable / an instance of outcome\\
$Y(\va_t, b, \vx_t)$ & \vtop{\hbox{\strut A potential outcome under baseline $b$,} \hbox{\strut sequential treatments $\va_t$ and time-varying}
\hbox{\strut covariates $\vx_t$ up to time $t$}}\\
\bottomrule[1pt]
\end{tabular}}
\label{tab:symbol}
\end{table}

\noindent \textbf{Time-varying confounding.} Given an \textit{observational} data $\dataD = \big\{(b^{(i)}, \vx_t^{\, (i)}, \va_t^{\, (i)}, y^{(i)})\big\}_{i=1}^N$, we observe a patient's baseline $b^{(i)}$, time-dependent covariates $\vx_t^{\, (i)}$ and treatment sequence $\va_t^{\, (i)}$ up to time $t$, and the final outcome $y^{(i)}$. We assume the observed outcome is \textit{consistent} with the potential outcome so that $y^{(i)} \equiv Y\big(\vA = \va_t^{\, (i)}, B = b^{(i)}, \vX = \vx_t^{\, (i)}\big)$. In the literature of machine learning, outcome predictions usually take  $\{(b^{(i)},\vx_t^{\, (i)},\va_t^{\, (i)})\}_{i=1}^N$ as input features, and fit regressions (if $\Y$ is continuous) or classifications (if $\Y$ is discrete) over the features in a supervised way such that the loss between predicted outcomes and factual outcomes $\{y^{(i)}\}_{i=1}^N$ are minimized. However, in the task of treatment outcome prediction, which requires also predictions on counterfactual outcomes. Otherwise these predictive methods without handling counterfactual outcomes lead to biased results. 
\begin{figure}[h]
    \centering
    \includegraphics[width= .2\textwidth]{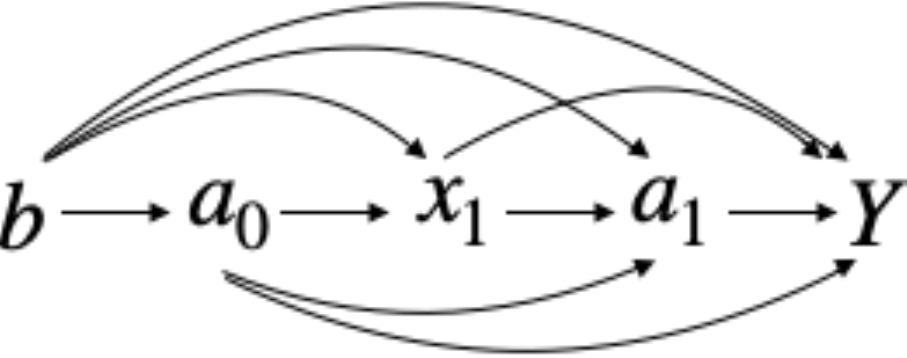}
    \caption{An illustration of time-varying confounding: $x_1$ is a confounder for later treatment $a_1$ and outcome $Y$, but is also affected by earlier treatment $a_0$ and baseline variable $b$.}
    \label{fig:varying}
\end{figure}

The bias occurs in these predictive models because treatment assignments usually were not \textit{randomized} in observational data: the decision of assigning $a_t$ at time $t$ usually depends on the history of covariates $\vx_t$ and past treatment assignments $\va_{t-1}$, whereas the covariates at next time step $x_{t+1}$ will also be changed by the new treatment assignment $a_t$. This is known as \textit{time-varying confounding}, as shown in Figure \ref{fig:varying}. For example, invasive treatments may only be assigned to those critically ill patients; and multiple treatments are also more likely given to those severe patients. In this paper, we aim to remove such time-varying bias from longitudinal observational data with the assumption of \textit{no unmeasured confounding}.
In continuous monitoring setting such as ICU, this assumption is not too strong as rich sets of measurements are continuously recorded.
This is a key assumption for being identifiable in causal inference, which assumes that all the factors affecting the decisions and outcomes are measurable by variables $B$ and $\vX$.

We illustrate the time-varying confounding in a short sequence of length two in Figure \ref{fig:varying}. We can expand and write the joint distribution of $(b, \vx_t, \va_t)$ at any given time $t$ as
\begin{multline*}
   \Prob(b, \vx_t, \va_t) = \Prob(b) \Prob(a_0\mid b) \cdot \\ \prod_{m = 1}^t \Pr(x_m \mid \va_{(m-1)},b, \vx_{(m-1)}) \Prob(a_m \mid \va_{(m-1)},b, \vx_m), 
\end{multline*}
where $\va_{(m-1)} = \{a_0,...,a_{m-1}\}$ and $\vx_{(m-1)} = \{x_1,..., \allowbreak x_{m-1}\}$. Time-varying confounding is presented in the coupled dependencies of $x_m$ on $\{\va_{(m-1)}, b, \vx_{(m-1)}\}$, and $a_m$ on $\{\va_{(m-1)}, b, \vx_m\}$ at each time step $m$. As discussed earlier, there are three main categories of bias adjustments on time-varying confounding. The IPTW methods focus on estimating the propensity scores $\Prob(a_m \mid \va_{(m-1)},b, \vx_m)$, while the g-formula methods consider this probability as fixed and focus on modeling $\Pr(x_m \mid \va_{(m-1)},b, \vx_{(m-1)})$ and $\E\big(Y(\va_t, b, \vx_t) \mid \vA = \va_t, B = b, \vX = \vx_t\big)$, and the doubly robust methods combining both so that only one of the models need to be correctly specified. 

\setlength{\textfloatsep}{0.1cm}
\begin{algorithm2e}[t]
\DontPrintSemicolon
\tcc{Recurrent weights generation}
\KwInput{$\bh= \bm{0}$, $n=0$, $N_1$, $\{\log \hat{\Prob}(A_m | \vA_{(m-1)})\}_{m=1}^t$}
\KwData{$\big\{(b^{(i)}, \vx_t^{\, (i)}, \va_t^{\, (i)})\big\}_{i=1}^N$}
   \While{$n < N_1$}
   {
        $\bh, \mathcal{M}_{ip} \leftarrow \arg \min_{\bh, \mathcal{M}_{ip}} \, \Loss_{IP}$.\\
   }
   Set $W_s^{(i)} = \exp \big( \sum_{m=1}^t \log \hat{\Prob} (a_m^{(i)} | \va^{(i)}_{(m-1)}) - \log \allowbreak \Prob_{\mathcal{M}_{ip}} (a_m^{(i)} |\bh_{(m-1)}) \big)$ for all $i$'s.\\
\KwOutput{$\{W^{(i)}_s\}_{i=1}^N$}
\tcc{===================================}
\tcc{Weighted outcome progression}
\KwInput{$\bm{g}= \bm{0}$, $n=0$, $N_2$, $\{W^{(i)}_s\}_{i=1}^N$}
\KwData{$\dataD$}
   \While{$n < N_2$}
   {
        $\bm{g}, \mathcal{M}_y \leftarrow \arg \min_{\bh, \mathcal{M}_y}\, \Loss_Y$.\;
   }
\KwOutput{$\bm{g}^*, \mathcal{M}^*_y$ }
\caption{\mname: Deep Recurrent Inverse TreatmEnt weighting}
\label{alg}
\end{algorithm2e}
\setlength{\floatsep}{0.1cm}

\subsection{Recurrent inverse treatment weighting} In this paper, we propose a {\it doubly robust} method that uses a recurrent neural network for recurrently encoding the history of past treatments and time-dependent covariates in the purpose of generating propensity scores, constructs weights per sequence based on the scores and fit the weighted sequences into another recurrent network for predicting the final potential outcomes. 

\noindent \textbf{\underline{Phase I}. Recurrent weights generation:} Instead of generating standard IPT weights, we construct the stabilized IPT weights as follows:
\begin{equation}
W_s = \frac{\prod_{m=1}^t \Prob\big( a_m \mid \va_{(m-1)}\big)}{\prod_{m=1}^t \Prob \big(a_m \mid \va_{(m-1)}, b, \vx_{(m-1)}\big)}.
\label{eq:IPTW}
\end{equation}
The standard IPT weights are now stabilized by multiplying the unconditional probability of treatment assignments, which can reduce the variance of the generated weights and also preserve the original cohort size when using them to re-weight the data. In this paper, we approximate the numerators by the empirical conditional probabilities. That is,
$$\hat{\Prob} \big( a_m \mid \va_{(m-1)}\big) = \frac{\text{\# of } \{\va_{(m-1)}, a_m\} \text{ observed in } \dataD \text{ by } m}{\text{\# of } \va_{(m-1)} \text{ observed in } \dataD \text{ by } m-1}.$$
To compute the denominator, we learn a sequential recurrent encoding $\bh_m = h(\vx_m, \va_m)$ at each time step $m = 1, 2, ..., t$, and fit a Logistic regression model to emit time-varying propensity scores (i.e., the denominator of E.q \ref{eq:IPTW}). 
We define the Logistic regression model as $\mathcal{M}_{ip}$:
$$\mathcal{M}_{ip}:~ a_m \sim \text{Sigmoid} \big(\textbf{w}_h^\top \bh_{(m-1)} + \textbf{w}_{bh}^\top b +  c_h \big).$$
Thus we can write the loss $\Loss_{\text{IP}}$ for the first network as:
\begin{align*}
  \Loss_{\text{IP}} = - \sum_{i=1}^N \sum_{m=1}^t \sum_{a \in \A} \mathbb{I}(a_m^{(i)} = a) \cdot \log \Prob_{\mathcal{M}_{ip}}\big(a_m^{(i)} = a\big)
  \label{eq:Loss1}
\end{align*}

\noindent \textbf{\underline{Phase II}. Weighted outcome progression:} Subsequently, we compute the weights $W_s$ and learn another recurrent encoding $\bm{g}_t = g(\vx_t, \va_t)$ for predicting the final outcome in a weighted regression model:
$$\mathcal{M}_y:~ W_s Y \sim f \big(\textbf{w}_{gy}^\top \bm{g}_{t} + \textbf{w}_{by}^\top b + \textbf{w}_{ay}^\top a_t +  c_y \big),$$
where $f$ is a Sigmoid function if the outcome variable $Y$ is categorical, and an identity function if $Y$ is real-valued. We can write the loss $\Loss_{Y}$ for the second network as follows:
\begin{equation*}
  \Loss_{Y} = \sum_{i = 1}^N  \hat{W}_s^{(i)} \cdot \Loss(y^{(i)}, \E_{\mathcal{M}_y}(Y^{(i)})),
  \label{eq:Loss2}
\end{equation*}  
where $W_s^{(i)}$'s are the estimated weights computed from Eq. \ref{eq:IPTW}. The loss function $\Loss$ is a binary cross entropy if $Y$ is categorical and mean squared error if $Y$ is real-valued. 

We summarize our two-phase pipeline, named \mname (Deep Recurrent Inverse TreatmEnt), in Algorithm \ref{alg}, where the goal of the first step is to solve $(\bh^*, \mathcal{M}^*_{ip}) = \arg \min_{\bh, \mathcal{M}_{ip}}\, \Loss_{\text{IP}}$ and the goal of the second step is to solve $(\bm{g}^*, \mathcal{M}^*_{y}) = \arg \min_{\bm{g}, \mathcal{M}_{y}}\, \Loss_{\text{Y}}$ respectively. Our pipeline incorporates recurrent neural networks into the standard doubly robust procedure of fitting weighted outcome regressions using the generated propensity-based weights, and enhance it to the extent of accommodating much larger and higher dimensional continuous data.    

\subsection{Marginal structural models for assessing treatment effects in longitudinal studies}
By using \mname, we are able to obtain unbiased predictions on potential outcome $\hat{Y}(\va, b, \vx)$ given any past treatment assignment $\va$, baseline variable $b$ and time-varying trajectory $\vx$. We can extend \mname and combine it with \textit{Marginal Structure Models} (MSMs) \cite{hernan2000marginal} to utilize the predicted IPT weights and counterfactuals for estimating marginal treatment effects varying over time. 

We briefly introduce two MSMs in this paper for estimating the longitudinal treatment effects at both population level and pre-specified group levels, which will be used later for evaluating counterfactual predictions on our real data. We refer to \cite{hernan2000marginal} for more details.

\noindent \textbf{Assess time-varying average treatment effects (ATEs).}
Taking an example of binary outcome variables, we can define the following linear MSM model for assessing the time-varying ATEs:
\begin{equation}
    \text{logit } \Prob \big(Y(\va_t, b ) = 1\big) = \beta_0 \cdot m + \beta_m \cdot a_m +  \beta_b^{\top} h(b),
    \label{eq:msm}
\end{equation}
where $a_m = 1$ if there was a treatment at time $m$ and 0 otherwise. The coefficient $\beta_m$, or strictly saying odds ratio $\exp(\beta_m)$, can be interpreted as how much the odds of $Y$ would have flipped have the treatment given at time $m$ vs. never given any treatments in the observation window. Note that the time-varying covariate $\vx$ no longer exists in the formula because it is considered to be independent from the treatment assignments after weighting; so it can be marginalized out when assessing the effect of treatments.    

To estimate parameters $\bm{\beta}$ in the above MSM model, we take the likelihood of potential outcomes $\hat{\Prob} \big(Y(\va_t, b)$ predicted from Phase II, and fit a weighted linear regression over them with the weights generated from Phase I.  

\noindent \textbf{Assess heterogeneous treatment effects (HTEs).}
For assessing HTEs, we can modify the MSM model to the following equation \cite{hernan2000marginal}:
\begin{multline}
    \text{logit } \Prob \big(Y(a, I_g, b ) = 1\big) = \\ \beta_0 \cdot m + \beta_a \cdot a + \beta_g \cdot I_g + \beta_{a_g} \cdot a I_g  + \beta_b^{\top} h(b),
    \label{eq:cmsm}
\end{multline}
where $a = 1$ if there was a treatment at time $m$ and 0 otherwise, $I_g = 1$ if an individual belongs to the pre-specified group $g$. Then the conditional odds ratio given group $g$ can be computed as $\exp(\beta_a + \beta_{a_g})$, which measures the effectiveness of treatment within the group.

\section{Experiments}
\label{sec:exp}
To evaluate \mname, we first conduct a simulation in which we know the ground truth, and then demonstrate its performance on a complex real-world data. 


\subsection{Simulations}
Inspired by the study in \cite{hill2011bayesian}, we design the following simulation by extending their static settings into longitudinal. In this study, we know the ground truth of the time-varying average treatment effects (ATEs) and can guarantee that the assumption of no unmeasured confounding has been satisfied.   

We start with a baseline $b$ of $20$ pre-treatment variables that are generated from a multivariate normal distribution with zero mean and random covariance matrix. Then we obtain time-varying variables $x_t$ by gradually decreasing the value of each variable with a fixed rate for no treatments, and instantly increasing the subsequent values by $C - t$ for treatment initiated at time $t$. We generate the final outcome $Y$ at the end of time $T$ from $N \big(\beta_b^\top b + \beta^\top x_T, 1\big)$, where the coefficients in vectors $\beta_b$ and $\beta$ are randomly picked from $[0, 1,2,3,4]$ with probabilities $(.3, .25, .2, .15, .1)$. Vector $\beta$ is then normalized (i.e., $||\beta||_2 = 1$) so that ATE in the randomized setting decreases harmoniously when treatment initiation time increases:
$$ \text{ATE}_t = E[Y(a = t) - Y(a \equiv \emptyset)] = C - t.$$

\begin{figure*}[t]
    \centering
    \includegraphics[width= .75\textwidth]{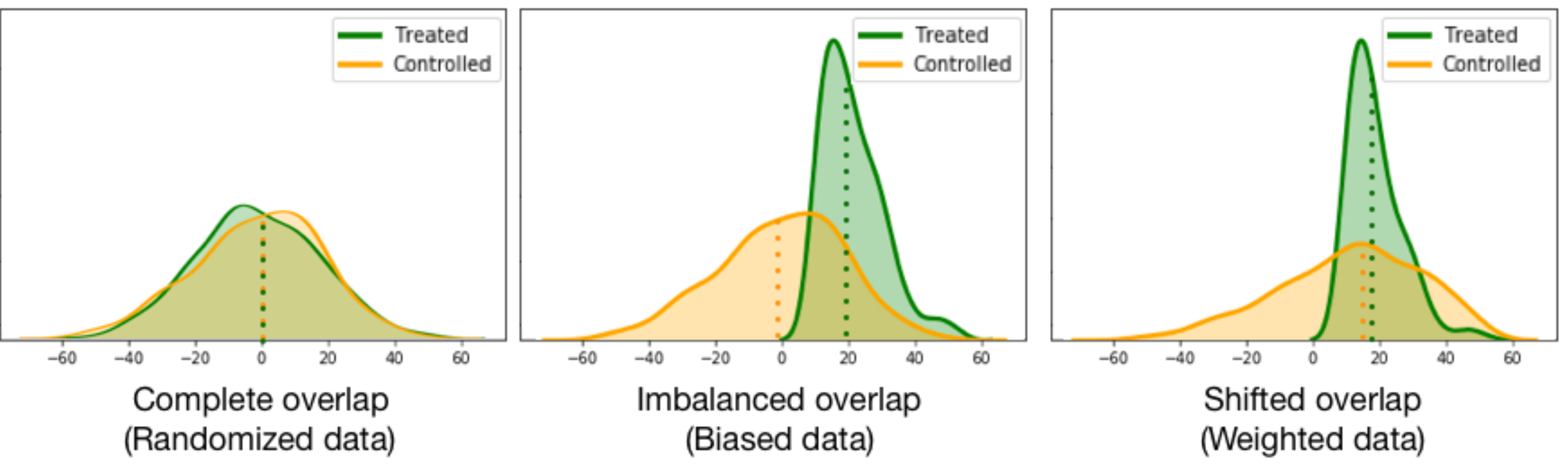}
    \caption{Empirical density plots of the baseline variable in the simulated data. \textbf{Left}: Densities in the treated and control are completely overlapped in the randomized data so estimation $\hat{\text{ATE}}$ is unbiased; \textbf{Middle}: Densities become lack of overlap after removing samples from the treated, $\hat{\text{ATE}}$ becomes biased since the means of the two groups are no longer matching due to the non-overlapped samples in the control; \textbf{Right}: Densities are shifted after weighting by \mname, the adjusted $\hat{\text{ATE}}$ is now unbiased because the means over the weighted samples are matching again.}
    \label{fig:sim_data}
\end{figure*}

We first simulate a randomized data containing $10,000$ samples, in which treatments are randomly initiated at time 1, 2,..., or $T$, or never initiated ($\emptyset$), with equal probabilities of $1/(T+1)$. This enables complete overlap between the control and treatment groups as shown on the Left of Figure \ref{fig:sim_data}. Then we create biased data by discarding samples in the treated while preserve all the samples in the control so distributions between the two groups are imbalanced as shown in the Middle of Figure \ref{fig:sim_data}. In details, we discard samples in two steps: a) remove the treated samples having $\beta_b^\top b < \lambda$ so distributions of baseline $b$ become partially overlapped; b) remove the remaining treated samples with probability of $(t-1)/\rho \cdot T$ given their treatment initiation time $t$ so distributions of $x_t$ become less overlapped over time. At the end we result at a data containing both time-invariant bias and time-varying bias, in which the level of bias (non-overlap) can be controlled by parameters $\lambda$ and $\rho$ respectively.  

\noindent \textbf{Results.} We fit Phase I of \mname for learning recurrent weights $W_s$ in Eq. \ref{eq:IPTW} on the simulated biased data. We estimate the ATE$_t$'s by taking the difference between two sample means, either adjusted with the weights or not, computed over the outcomes in the treated (treatment initiated at time $t$) and in the control. We compute root mean squared errors (RMSEs) of the estimations with respect to the truth ATE values. Here we pick $T = 3$ and constant $C = 4$, so the ground truth of ATE is $[3,2,1]$ indexing by treatment initiation time $t$. 

As we show in the middle of Figure \ref{fig:sim_data}, in which the support of $b$ is reduced to 1-D by the transformation of $\beta_b^\top b$, treated samples are removed by setting $\lambda = 0$ and $\rho = 1$ so distributions over the treated and controlled samples are imbalanced. With no adjustment on the bias, the RMSE of the ATE empirical estimation is high as $15.0$. By weighting the samples via \mname, the two distributions become shifted and their means become overlapped again, so the RMSE of the adjusted ATE estimation reduces to $0.44$ whereas the RMSE of the empirical estimation from randomized data is $0.21$. 

\begin{figure}[h]
    \centering
    \includegraphics[width= .49\textwidth]{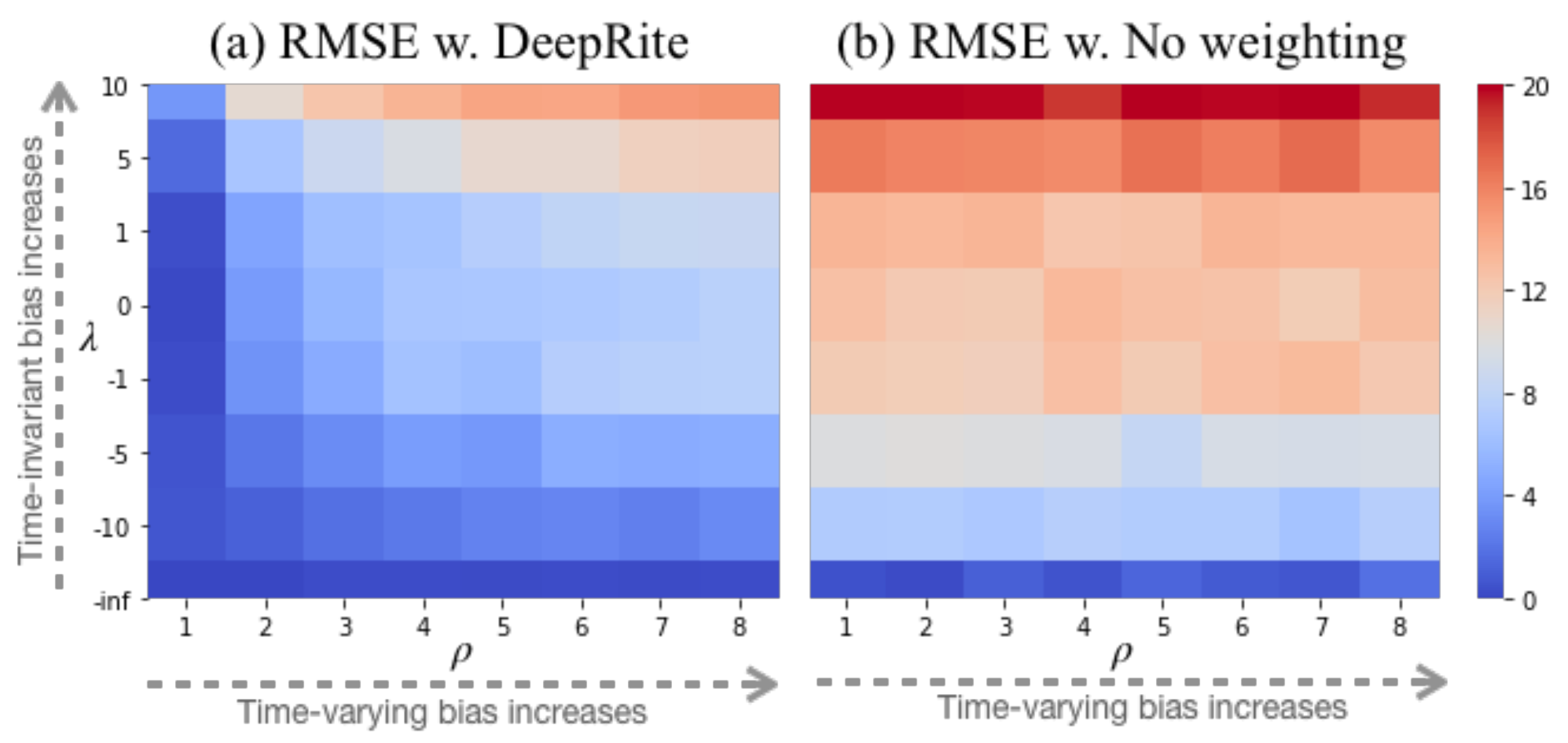}
    \caption{
    Heatmaps of RMSEs in estimating ATE by increasing time-varying bias (x-axis) and time-invariant bias (y-axis) in the simulated data. (a) RMSEs of \mname increases slowly from $0.20$ (lower left) to $13.1$ (upper right); (b) RMSEs of no weighting increases drastically from $0.21$ to $21.5$.}
    \label{fig:sim_rmse}
\end{figure}

We also vary the level of bias in the simulated data by increasing $\lambda$ (larger time-invariant bias) and $\rho$ (larger time-varying bias), and show in Figure \ref{fig:sim_rmse} that RMSEs of \mname increases slowly from $0.20$ to $13.1$ as $\lambda$ goes from -inf to 10 and $\rho$ goes from 1 to 8, while RMSEs of no weighting increases drastically from $0.21$ to $21.5$.

\subsection{Real data experiments}
Now we assess \mname on more complex but useful real-world data. It comes from the MIMIC-III Critical Care Database \footnote{\url{https://mimic.physionet.org/about/mimic/}} \cite{johnson2016mimic}.
The Database consists of deidentified health records from over $40,000$ critically ill patients who stayed in the intensive care units (ICUs) of the Beth Israel Deaconess Medical Center between 2001 and 2012. We study on a cohort of $19,954$ adult sepsis patients. Sepsis and septic shock are considered as one leading cause of mortality and critical illness in ICUs \cite{fleischmann2016assessment}. Our ultimate goal is to analyze, given any initiation time of fluid resuscitation to the cohort, to what extent the average risk of developing septic shock could be reduced. Furthermore, we also target at 5 other outcomes that are also important for monitoring during sepsis management; those are, whether a vasopressor is needed to maintain a mean arterial pressure (MAP) $\geq$ 65 mm Hg, serum lactate level $>$ 2 mmol/L, urine output $\geq$ 5 ml/kg/hr, SvO2 $\geq 70\%$ and CVP of $8 - 12$ mmHg. 
Details about cohort selection and MIMIC data preparation are described in Appendix A\footnote{Source code is provided in the supplementary material.}. In summary, the data contains 8 baseline variables, 35 time-varying covariates, and 6 binary outcomes that require multi-task classification in Phase II.  

We first evaluate \mname on factual predictions and compare its performance with purely predictive models. Then we validate counterfactual predictions by combining MSM and evaluating the estimated treatment effects at both population level and pre-specified subgroup levels.



\subsubsection{Experiment 1: Evaluations on factual predictions}
In this experiment, we take each patient's 12-hour observed data starting from their onset of sepsis, and predict at the end whether the patient will develop septic shock or fail any of the 5 targets in the next 6 hours. We train our models on $75\%$ of the cohort and hold out $25\%$ for testing. 

\noindent \textbf{Experimental setup.}
To fit outcome progression, we fit a 
bidirectional Long Short-Term Memory (LSTM) network with 2 layers and hidden size of $300$ for encoding the histories of the past observed measurements and treatments. We combine the baseline features $b$ with the last hidden state and perform 2 MLP layers with $0.2$ dropout before the final Sigmoid activation for the multi-class classification. We use Adam optimizer, and choose a batch size of 128 and learning rate of $1$e$-3$. We use this network as one baseline as it predicts outcomes without taking into account the relationship between treatments and outcomes. By demonstrating the prediction power, along with the ease of auto encoding on sequences, of the recurrent neural network, we also pick a linear classifier SGD and an ensemble classifier AdaBoost as two other baselines. We handcraft historical features by taking min, average and max of the $29$ real-valued time-varying variables, combine them with their current values plus baseline features and eventually reach at a total of $130$ features.  

To adjust time-varying confounding, we fit \mname by first training one LSTM for generating the stabilized IPT weights and then the above LSTM for weighted multi-task progression. In IPT weight generations, we use the same network structure for encoding the histories, and combine baseline features $b$ with the hidden state at each time step for emitting the time-varying probability of initiating the resuscitation at next step. As a comparison, we also apply the traditional IPTW method for generating weights, which is to sequentially fit Logistic regression over the features we handcraft at each time step. Then we use these weights to fit weighted SGD and AdaBoost respectively and take them as two other baselines.

Further, one disadvantage of IPT weighting is that large weights can emerge and outcome predictions can become unstable. For example, a patient who was very unlikely to be treated but ended up being treated will receive an extreme high weight. This happens because the emitted probability of being treated may result at extreme values, either too low as 0.0 or too high as 1.0. Previous methods ~\cite{cole2008constructing} truncate the inverse weights at $(.01, .99)$ or $(.05, .95)$ quantiles. Alternatively, we posit an $L_2$ regularization on the regression weights $\bm{w}$ in $\mathcal{M}_{ip}$ that automatically smooths the estimated propensity scores and obtains non-extreme inverse weights. We call this Smoothed model as \mnameS. To summarize, we compare the following models in this experiment: 
 \begin{itemize}
    \item SGD:  A linear classifier on handcrafted features.
    \item AdaBoost: An ensemble classifier on handcrafted features.
     \item LSTM: A vanilla predictive LSTM.
     \item IPTW-SGD: SGD using traditional IPT weights.
     \item IPTW-AdaBoost: AdaBoost using traditional IPT weights.
     \item \mname (Proposed): LSTM using recurrent weights.
     \item \mnameS (Proposed): Smoothed \mname.
\end{itemize}

\noindent \textbf{Results.} Before reporting the final prediction performance, we first compare the efficiency of training \mname in a pipeline (i.e., training two LSTMs separatedly) vs. end-to-end (i.e., training only one LSTM whose hidden states are shared in prediction of propensity scores and outcomes). Training two networks in a pipeline is found to be more efficient than training one network end-to-end: it takes around $2,000$ epochs for the pipeline to reach $90\%$ AUC-ROC score for predicting septic shock on the test set, whereas it takes about $7,000$ ($3.5 \times$ slower) epochs for the end-to-end network to reach $89\%$ AUC-ROC. More details about the comparison are described in Appendix. B. 

Additionally, we also compare the estimated propensity scores and stablized IPT weights by the non-smoothed model \mname vs. the smoothed model \mnameS in Table \ref{tab:iptw}. We see that the $L_2$ regularization imposes smoothness in the propensity scores and generate more stable reverse weights with their maximum bounded and mean near 1.0.   
\begin{table}[h]
\centering
\caption{Estimation of stabilized IPT weights}
\resizebox{\columnwidth}{!}{
\begin{tabular}{lllllll}
\toprule
 & \multicolumn{3}{c}{Propensity score} & \multicolumn{3}{c}{Stablized IPT Weight} \\
  \midrule
& Q$_{.01}$ & Avg. & Q$_{.99}$ & Min & Avg. & Max \\
 \cmidrule(r){2-4} \cmidrule(r){5-7}
No Smooth & .00 & .01 & .99 & 0.02 & 0.56 & 7.59 \\
$L_2$-Smooth & .04 & .05 & .62 & 0.02 & 0.92 & 3.88\\
\bottomrule
\end{tabular}}
\label{tab:iptw}
\end{table}

Now we report the prediction performance of all the models in Table \ref{tab:pred}. We tune \mnameS by varying the coefficient of $L_2$ regularizer from $5$e$-2$, $5$e$-1$, $1$, $5$ to $10$, and choose the best model (with coefficient $1$) to report. We see that \mnameS preserves equivalent prediction performance (mostly) as the predictive LSTM, both of which significantly outperforms traditional ML methods. Next we show that, better than LSTM, \mnameS advances in making accurate counterfactual predictions in terms of producing interpretable causal effect estimations.
\begin{table}
\centering
\caption{AUC-ROC scores ($\%$) of multi-task classification}
\resizebox{\columnwidth}{!}{
\begin{tabular}{lllllll}
\toprule
 & Shock & CVP & Lactate & SvO2 & U/O & Vasso \\
 \toprule
SGD & 87.7 & 91.7 & 94.5 & 95.5 & \textbf{69.8} & 79.7 \\
AdaBoost & 87.9 & \textbf{93.2} & 94.3 & 97.8 & 69.6 & 83.2 \\
LSTM & \textbf{90.0} & 92.6 & \textbf{95.5} & \textbf{98.0} & \textbf{69.8} & \textbf{86.0} \\
\midrule
IPTW-SGD & 87.9 & 91.9 & 93.9 & 95.8 & \textbf{69.8} & 81.4 \\
IPTW-AdaBoost & 86.6 & 92.9 & 94.1 & 97.8 & 68.7 & 81.5 \\
\mname & 89.1 & 92.4 & \textbf{95.5} & 97.6 & 68.6 & 82.1 \\
\mnameS & \textbf{90.0} & 92.6 & \textbf{95.5} & 97.2 & 69.5 & 84.7\\
\bottomrule
\end{tabular}}
\label{tab:pred}
\end{table}

\subsubsection{Experiment 2: Assessments on longitudinal treatment effect.}
 In this experiment, 
 our goal is to estimate the timing effect of initiations of fluid resuscitation on the development of septic shock. By combining MSM models, we formulate this problem as estimating the time-dependent odds ratio of developing septic shock given fluid resuscitation initiated at 0, 1, and up to 11 hours since sepsis is recognized as opposed to no resuscitation given in the first 12 hours. 
 
\begin{figure*}[t]
    \centering
    \includegraphics[width= .8\textwidth]{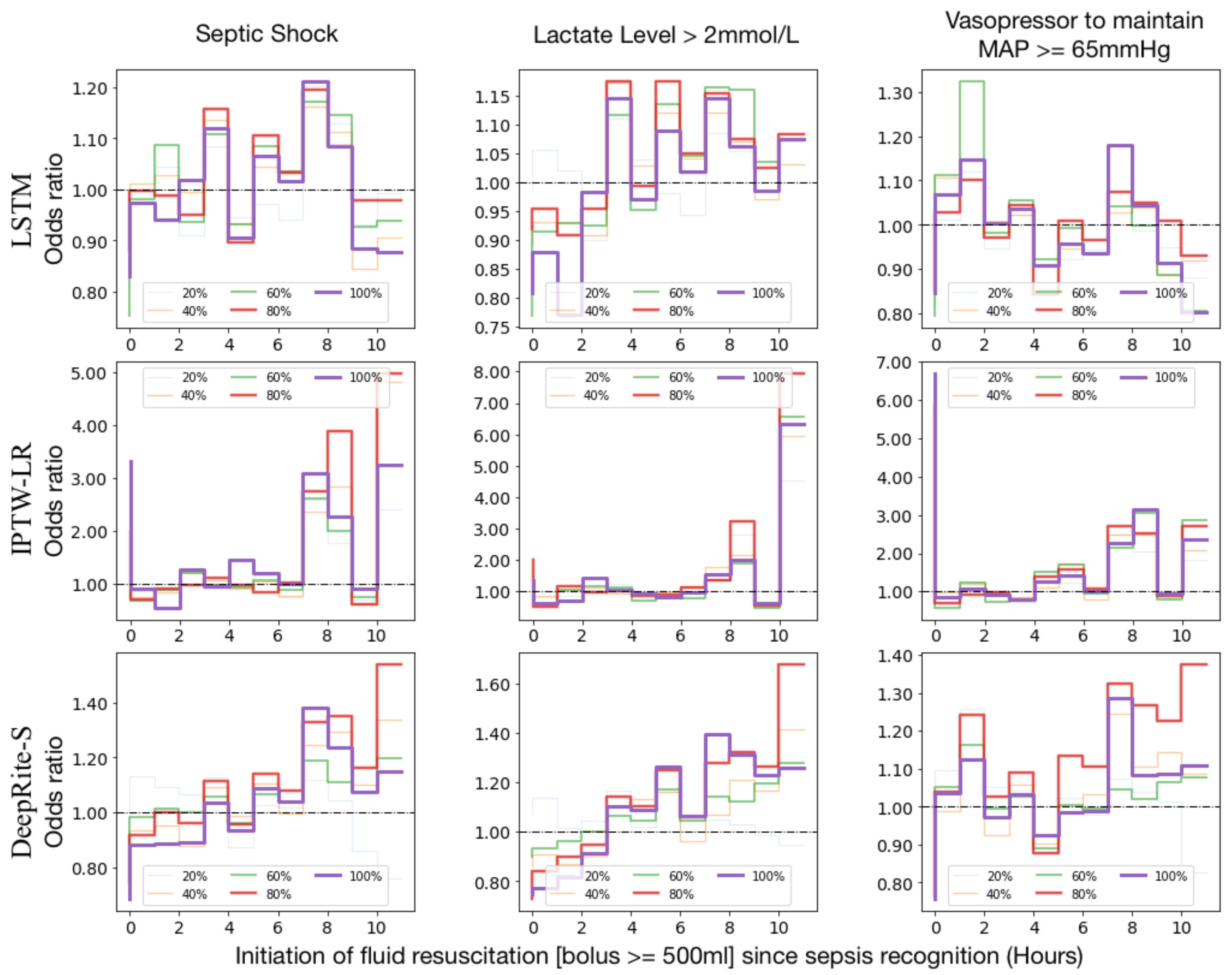}
    \caption{Step functions of the estimated hazard odds ratios over the fluid initiation time. Compared to the baseline models, \mnameS is able to align with the SCC guidelines such that hazards of developing septic shock consistently increases as the initiation of fluid resuscitation delays.
    }
    \label{fig:msm}
\end{figure*}

\noindent \textbf{Results on ATE estimations.}  
We fit \ref{eq:msm} on top of the predicted likelihood of septic shock, along with the other 2 outcomes that are used for defining it, from LSTM (with no weighting), IPTW-LR and \mnameS respectively.
Then we plot the estimated hazard odds ratios $\exp(\hat{\beta}_m)$'s as step functions over time $m$ in Figure \ref{fig:msm}. We say an estimation is good if the step function of the hazard ratio increases consistently as the time $m$ goes further. Because it would match the SSC Guidelines \cite{rhodes2017surviving} that recommends fluid resuscitation should be initiated as early as possible when sepsis is recognized in order to lower the risk of septic shock. From Figure \ref{fig:msm}, we see that only the ratios estimated from \mnameS consistently increase across all the three outcomes. Traditional causal model IPTW-LR is still able to capture the overall trend compared to LSTM regardless how well LSTM makes factual predictions. Furthermore, the guidelines also highly suggest that fluid resuscitation should happen in the first 6 hours of sepsis; this again aligns with our observation from Figure \ref{fig:msm} that the odds ratios for developing septic shock in the first 6 hours of fluid initiation are mostly below $1$ but exceeds the threshold consistently afterwards. The odds ratio estimation for all the 6 outcomes are reported in Appendix C.

In addition, we also evaluate the robustness of \mnameS to data imbalance in this real data. We mix our training data by taking different proportions (e.g. $20\%, 40\%, ..., 100\%$) of the treated data into the untreated data, and plot the corresponding ratio steps in different colors in Figure \ref{fig:msm}. We see that the estimations from \mnameS are more stable and is able to uniformly capture the increasing trend regardless of the level of imbalance.

\noindent \textbf{Results on HTE estimations.} From Result 1, we validate that the learnt average hazard ratios from \mnameS can match the findings from SCC Guidelines. Now we aim to assess the HTE estimations inferred by \mnameS comparing to LSTM with no weighting. We first group our cohort in two ways: 1) Group patients who eventually deceased in the ICU and who were sent to step-down alive; 2) Group patients whose fluid rates got tuned up after initiation and whose rates remain unchanged or tuned down after initiation. 
Note that these events are post hoc and they were never used as input features in our study, so we are able to use them for validating our HTE estimations. We fit \ref{eq:cmsm} over septic shock predictions, where the groups are specified by alive patients vs. deceased, or patients with fluid rate tuned up vs. not, and obtain the conditional hazard odds ratios within groups.

Table \ref{tab:cate} summarizes our results. The hazard of developing septic shock is expected to be higher in the deceased group than the alive group. Similarly, the hazard is expected to be higher in the patients whose fluid rate get tuned up comparing with fluid rate remains or got tuned down. That is because, higher the hazard ratio, lower the effectiveness of the current fluid treatment, therefore higher the fluid rate to be increased. From Table \ref{tab:cate}, we see \mnameS perfectly matches the expected order with   
Spearman's $\rho$ of $1.0$ in both analysis, whereas LSTM misses matching completely with $\rho$ of $-1.0$.
\begin{table}[h]
\caption{The estimated conditional hazard odds ratio of developing septic shock within groups. Estimations from \mnameS perfectly match the expected ranks that hazard is higher in Deceased patients comparing to Alive, and also higher in patients whose Fluid rate get tuned up comparing to not, whereas LSTM misses matching completely.}
\begin{tabular}{l|cc|r}
\toprule
 &  Alive  & Deceased & Rank $\rho$\\
\midrule
\mnameS & 0.705 & 0.825 & \textbf{1.0}\\
LSTM & 0.946 & 0.903 & -1.0\\
\bottomrule
\end{tabular}
\newline
\vspace*{.3em}
\newline
\begin{tabular}{l|cc|r}
\toprule
& Fluid rate  $\rightarrow \downarrow$ & Fluid rate $\uparrow$& Rank $\rho$\\
\midrule
\mnameS & 0.930 & 0.987 & \textbf{1.0} \\
LSTM & 0.793 & 0.750 & -1.0\\
\bottomrule
\end{tabular}
\label{tab:cate}
\end{table}


\section{Conclusions}
In this paper, we propose a DL pipeline \mname for efficiently removing treatment bias from the large and complex longitudinal observational data. We evaluate \mname on both synthetic data and a complex real-world health data. We show that it can recover the ground truth in the simulated bias data. Comparing to traditional ML methods and causal methods, we show its powerful factual predictions and accurate counterfactual inference on a complex high-dimensional dependent longitudinal observational data.


\bibliographystyle{named}
\bibliography{references}

\newpage
\section*{Appendix A. Data preparation for analyzing effect of fluid resuscitation on sepsis patients using MIMIC-III Database}
According to the definition of Sepsis-3 \cite{singer2016third}, the onset of sepsis is defined to be the time when an increase in the Sequential Organ Failure Assessment (SOFA) score of $2$ points or more occurs in response to infections. 
We use the \textit{Sepsis-3} toolkit\footnote{\url{https://doi.org/10.5281/zenodo.1256723}} to obtain the suspected infection time in patients, 
and following the process in \cite{seymour2016assessment} to identify the onset of sepsis. 
We result at a total of $20,009$ sepsis patients with age $\geq 18$ from MIMIC-III database. For the purpose of longitudinal studies in this paper, we exclude those patients who stay in ICUs less than $6$ hours since the onset of sepsis and obtain a final cohort of size $19,954$. 

\begin{figure}
    \centering
    \includegraphics[width= .5\textwidth]{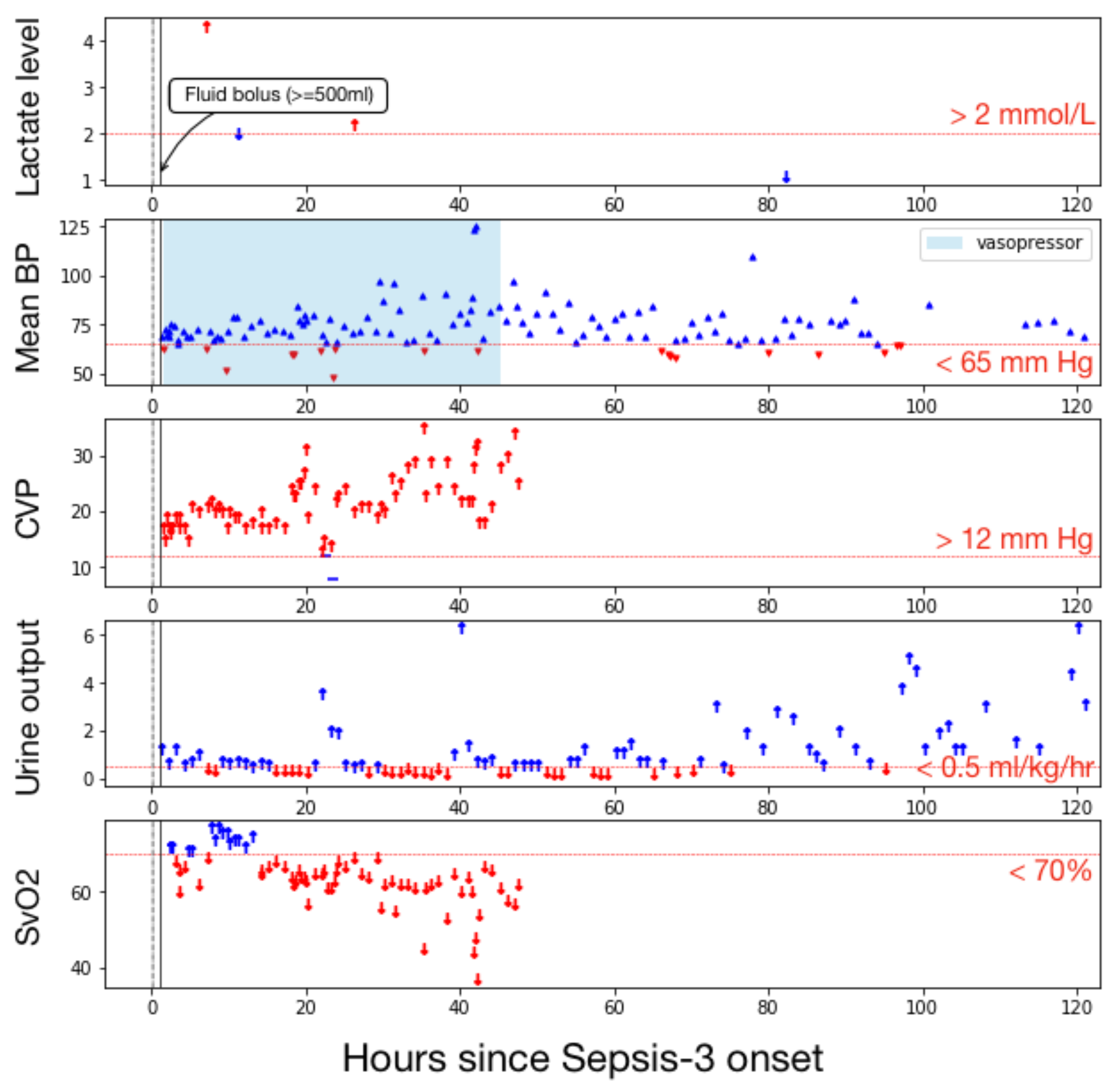}
    \caption{An example trajectories of 5 targeted outcomes in monitoring septic shock on sepsis patients.}
    \label{fig:demo}
\end{figure}

Intravenous fluids (only Crystalloids and Colloids are considered in this paper) are highly recommended in the early management of sepsis \cite{rhodes2017surviving}, and particularly fluid resuscitation of bolus $\geq 500$ mL is one of the most common treatment for managing septic shock. In our experiment, we consider two treatment types: initiation or no initiation of fluid resuscitation. We discover $8,135$ sepsis patients ($41\%$ the entire cohort) who have had fluid resuscitated, and $11,819$ sepsis patients having no fluid resuscitation during the observation window.
As one of the outcomes, septic shock can be identified by a vasopressor requirement to maintain a mean arterial pressure (MAP) $\geq 65$ mm Hg and serum lactate level $> 2$ mmol/L ($>18$ mg/dL) \cite{singer2016third}. In addition, three more targets are included as outcomes since they also need to be monitored during sepsis management: those are \textit{urine output} $\geq 5$ ml/kg/hr, \textit{SvO2} $\geq 70\%$, and CVP of $8-12$ mmHg. An example patient, whose 5 trajectories are under monitored simultaneously during sepsis management, is shown in Figure \ref{fig:demo}; the patient was diagnosed as septic shock as the first two variables met the criteria. Thus we have a problem of multi-task classification for predicting whether a sepsis patient would have developed septic shock or failed in the other 5 targets had them fluid resuscitated or not.
\begin{figure}
    \centering
    \includegraphics[width= .4\textwidth]{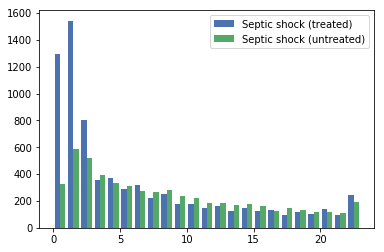}
    \caption{Number of sepsis patients in the cohort who developed septic shock or not with or with out fluid resuscitation.}
    \label{fig:stats}
\end{figure}

From Figure \ref{fig:stats}, we can also see that removing bias is important in this treatment outcome prediction task. In the figure, septic shock is observed to be developed significantly more in those patients who have had fluid administrated while less in those patients who haven't. Without taking into account that sicker patients are more likely being treated, one may mistakenly conclude that fluid resuscitation are more likely to lead to septic shock. This is also the motivation why we apply \mname on this data for bias adjustment. 

We construct baseline features $b$ by extracting the patient's age, gender, race, height, weight, sepsis onset hour since ICU admission, whether diagnosed diabetes or on a ventilator at ICU admission. We generate time-dependent features $x_t$ per hour, including 8 vital signs, 16 lab measurements, urine output, venous oxygen saturation (SvO2), central venous pressure (CVP), dosage and duration indicators of 6 vasopressors, and duration indicators of continuous renal replacement therapies (CRRT) and ventilation. The 8 vital signs include heart rate, systolic blood pressure, diastolic blood pressure, mean blood pressure, respiration rate, temperature, SpO2 and glucose; the 16 lab measurements include Anion gap, Albumin, Bands, Bicarbonate, Bilirubin, Creatinine, Chloride, Glucose, Hematocrit, Hemoglobin, Lactate, Platelet, Potassium, PTT, INR, PT, Sodium, BUN and WBC; the 6 vasopressors include dobutamine, dopamine, epinephrine, norepinephrine, phenylephrine, and vasopressin. We fill missing values like lab measurements using the last measured value; we clamp real-valued features in between their $0.05$-quantile and $0.95$-quantile values respectively and normalize the features using min-max normalization. 

\section*{Appendix B. Comparison of running-time efficiency between training two networks in a pipeline vs. training one network end to end}

\begin{figure}
    \centering
    \includegraphics[width= .4\textwidth]{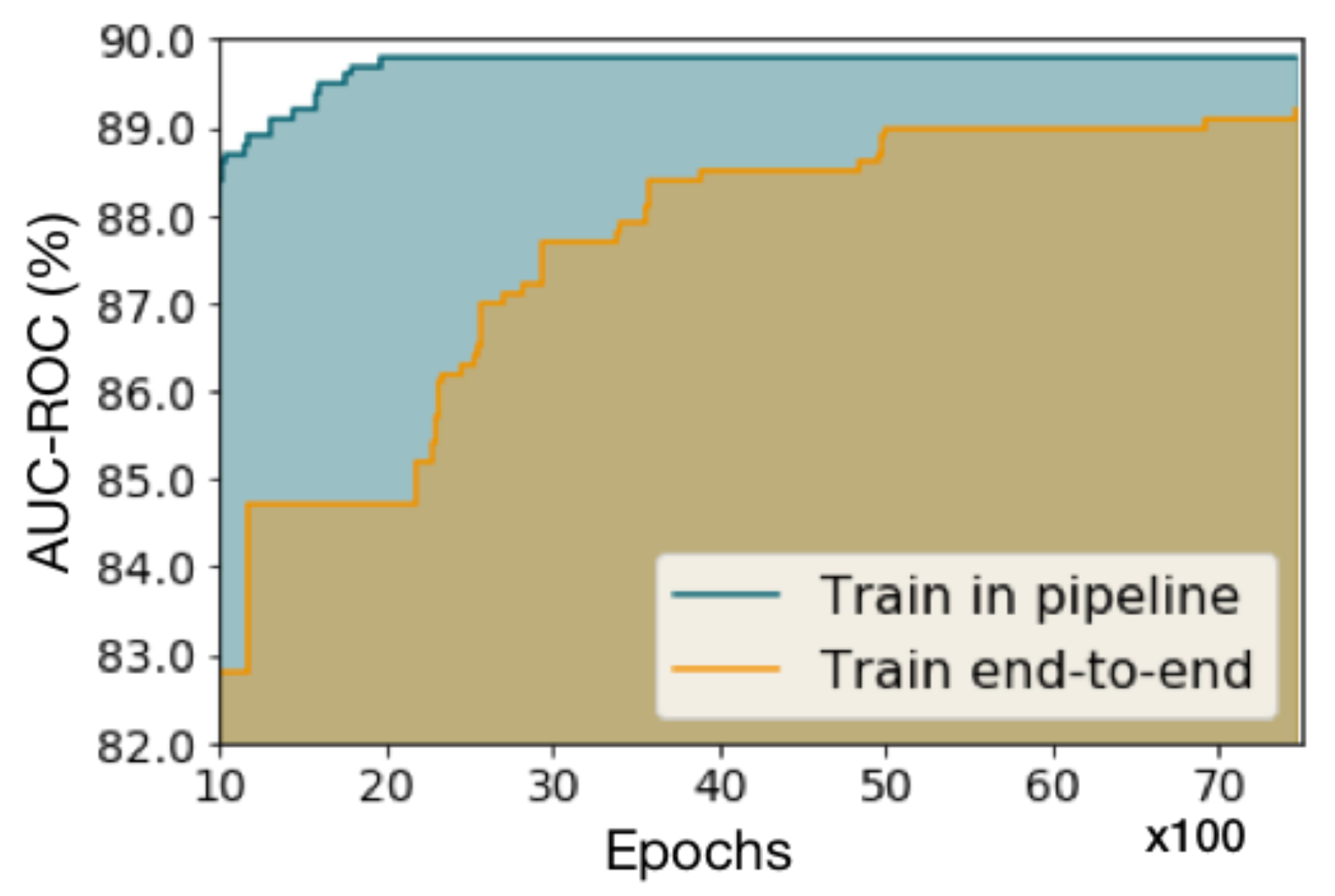}
    \caption{Convergence of prediction AUC-ROC by the number of epochs needed for training \mname in a pipeline vs. end to end.}
    \label{fig:conv}
\end{figure}

In Figure \ref{fig:conv}, we can see that it is more efficient to train a pipeline than train an end-to-end model. For training in a pipeline, we first train an LSTM on treatment prediction for 100 epochs and use the generated weights to fit another LSTM on outcome prediction. Here, one epoch means the model takes one batch of data for training. For training end to end, we train only one LSTM and alternately optimize between the $\Loss_{\text{IP}}$ and $\Loss_Y$. it takes around $2,000$ epochs for the pipeline to reach $90\%$ AUC-ROC score for predicting septic shock on the test set, whereas it takes about $7,000$ ($3.5 \times$ slower) epochs for the end-to-end network to reach $89\%$ AUC-ROC.

\section*{Appendix C. Odds ratio estimation for all the 6 outcomes using the marginal structural model}

Note that fluid resuscitation is expected to have positive effects on Septic shock, lactate level, onset of vasopressor and negative effects on CVP and SvO2. The effect on urine output is potentially negative, but is also confounded by other fluid inputs that are not included in our study.   
\begin{figure*}
    \centering
    \includegraphics[width= .8\textwidth]{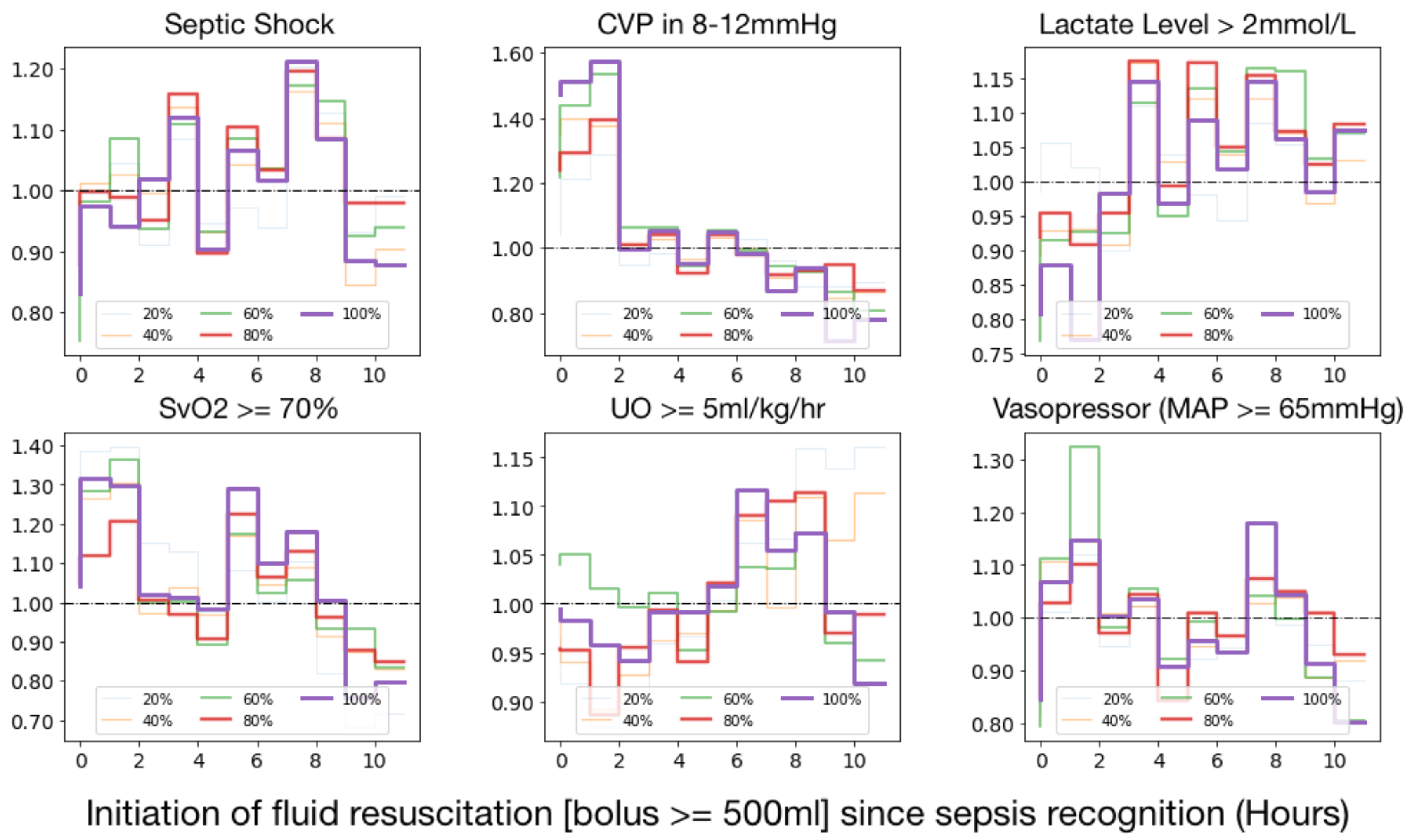}
    \caption{Odds ratio estimation for all the 6 outcomes based on LSTM.}
    \label{fig:stats}
\end{figure*}

\begin{figure*}
    \centering
    \includegraphics[width= .8\textwidth]{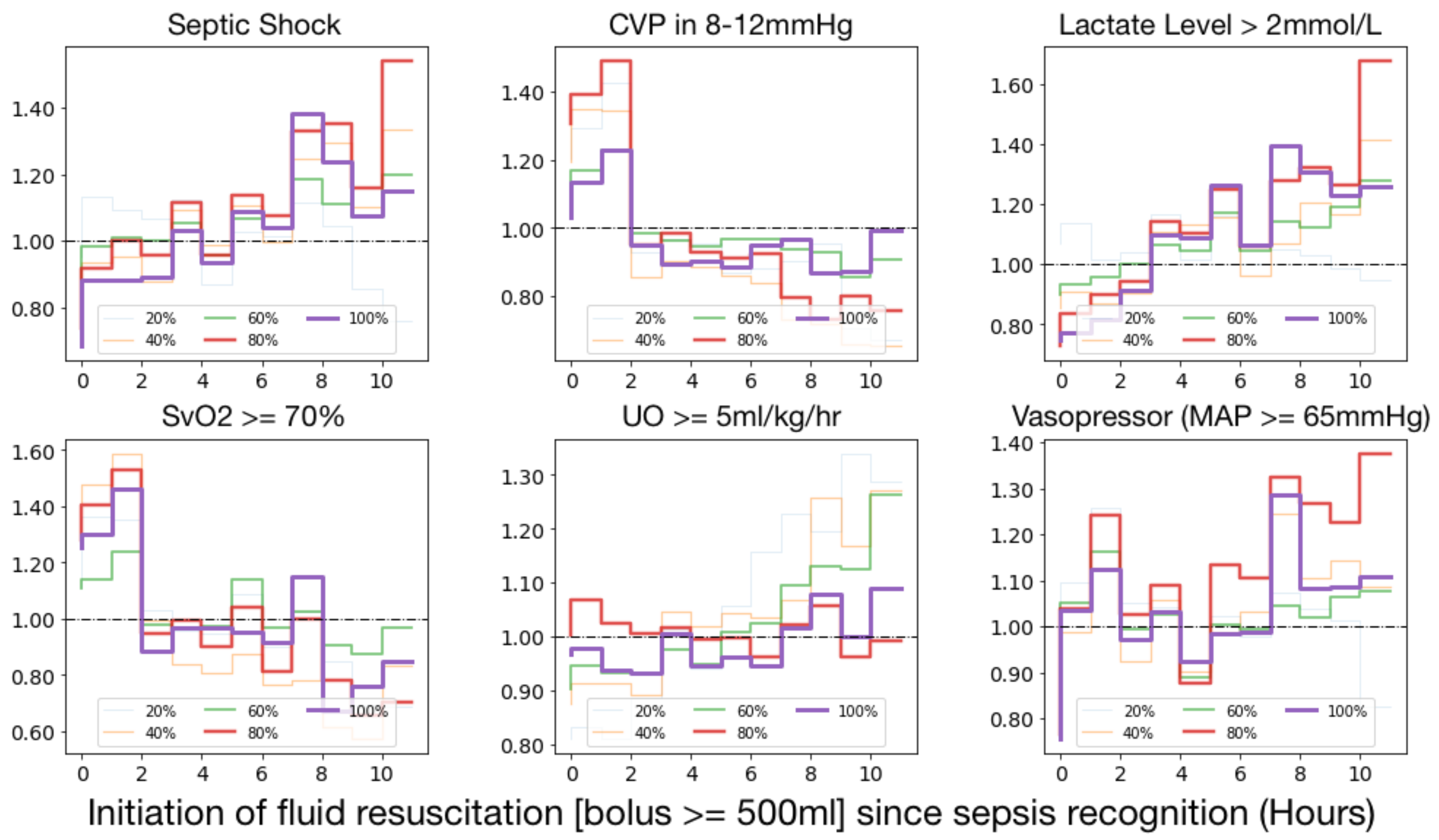}
    \caption{Odds ratio estimation for all the 6 outcomes based on \mnameS.}
    \label{fig:stats}
\end{figure*}
\end{document}